\documentclass{article}

\usepackage[preprint]{neurips_2023}

\usepackage[utf8]{inputenc} % allow utf-8 input
\usepackage[T1]{fontenc}    % use 8-bit T1 fonts
\usepackage{hyperref}       % hyperlinks
\usepackage{url}            % simple URL typesetting
\usepackage{booktabs}       % professional-quality tables
\usepackage{amsfonts}       % blackboard math symbols
\usepackage{nicefrac}       % compact symbols for 1/2, etc.
\usepackage{microtype}      % microtypography
\usepackage{xcolor}         % colors

\usepackage{graphicx}
\usepackage{amsmath}
\usepackage{amssymb}
\usepackage{bm}
\usepackage[accsupp]{axessibility}
\usepackage{natbib}
\setcitestyle{numbers,square}

\RequirePackage{xspace}
\makeatletter
\DeclareRobustCommand\onedot{\futurelet\@let@token\@onedot}
\def\@onedot{\ifx\@let@token.\else.\null\fi\xspace}

\def\ie{\emph{i.e}\onedot}

\def\etal{\emph{et al}\onedot}
\makeatother

\title{Learning 3D Geometry and Feature Consistent Gaussian Splatting for Object Removal}

\author{%
  Yuxin Wang \\ HKUST \\
  \And
  Qianyi Wu \\ Monash University \\
  \And
  Guofeng Zhang \\ Zhejiang University \\
  \And
  Dan Xu \\ HKUST \\
}

\begin{document}

\maketitle

\begin{abstract}
    This paper tackles the intricate challenge of object removal to update the radiance field using the 3D Gaussian Splatting~\cite{3dgs}. The main challenges of this task lie in the preservation of geometric consistency and the maintenance of texture coherence in the presence of the substantial discrete nature of Gaussian primitives. We introduce a robust framework specifically designed to overcome these obstacles. The key insight of our approach is the enhancement of information exchange among visible and invisible areas, facilitating content restoration in terms of both geometry and texture. Our methodology begins with optimizing the positioning of Gaussian primitives to improve geometric consistency across both removed and visible areas, guided by an online registration process informed by monocular depth estimation. Following this, we employ a novel feature propagation mechanism to bolster texture coherence, leveraging a cross-attention design that bridges sampling Gaussians from both uncertain and certain areas. This innovative approach significantly refines the texture coherence within the final radiance field. Extensive experiments validate that our method not only elevates the quality of novel view synthesis for scenes undergoing object removal but also showcases notable efficiency gains in training and rendering speeds. Project Page: \url{https://w-ted.github.io/publications/gscream}
\end{abstract}

% Introduction
\section{Introduction}
\label{sec:intro}

3D object removal from pre-captured scenes stands as a complex yet pivotal challenge in the realm of 3D vision, garnering significant attention in computer vision and graphics, particularly for its applications in virtual reality and content generation. This task extends beyond the scope of its 2D counterpart, \ie image in-painting~\cite{bertalmio2000image}, which primarily focuses on \emph{texture filling}. In 3D object removal, the intricacies of \emph{geometry completion} become equally crucial, and the choice of 3D representation plays a significant role in the effectiveness of the model and rendering quality.~\cite{kazhdan2006poisson,dai2017complete,dai2018scancomplete,dai2020sg,shih20203d,liu2022nerfin,wang2023inpaintnerf360}. 

\begin{figure}[!tp]
    \centering
    \includegraphics[width=\linewidth]{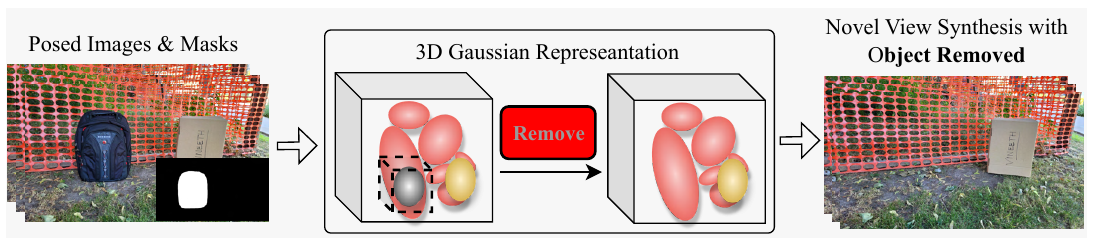}
    \caption{\textbf{Illustration of the Object Removal using 3D Gaussian Representations.} Given a set of multi-view posed images and object masks, our goal is to learn a 3D consistent Gaussian representation modeling the scene with the object removed, which enables the consistent novel view synthesis without the specific object. }
    \label{fig:teaser}
\end{figure}

Recently, the radiance field representation has revolutionized the community due to the superior quality of scene representation and novel view synthesis. Among these, the Neural Radiance Field (NeRF)~\cite{mildenhall2021nerf} has emerged as the groundbreaking implicit 3D representation approach, offering photo-realistic view synthesis quality. The high-quality rendering capabilities of NeRF have spurred further development in 3D object removal techniques based on it~\cite{liu2022nerfin,mirzaei2023spin,weder2023removing,yin2023or,mirzaei2023reference}. However, the intrinsic drawbacks of implicit representation, particularly its slow training and rendering speeds, pose severe limitations for practical applications based on object removal. For instance, it is highly expected that the system can quickly model the scene given any object mask condition for object removal, which enforces a straight requirement in terms of training efficiency. Another critical issue is that the object removal task relies on a flexible scene representation that can learn effective multi-view consistency to synthesize high-quality scene images with objects masked.

To effectively address the dual challenges of producing an enhanced radiance field for object removal, we introduce a pioneering strategy leveraging 3D Gaussian Splatting (3DGS)~\cite{3dgs}. Unlike implicit representations, 3DGS explicitly models the 3D scene using tons of Gaussian primitives. This approach has demonstrated notable advances in rendering efficiency and quality, surpassing traditional NeRF-based methods~\cite{muller2022instant,barron2022mip}. However, applying 3DGS to object removal presents unique challenges, primarily from two aspects: 1) \emph{Geometry Accuracy}: The inherently discrete nature of a significant number of Gaussians can result in an inaccurate representation of the underlying geometry in the standard 3DGS model. This inaccuracy poses a considerable challenge in executing geometry completion and ensuring geometric consistency in the object removal areas within a 3D space. 2) \emph{Texture Coherence}: Filling the region behind the removed object with consistent textures under the 3DGS framework represents another unexplored challenge. Achieving texture coherence across various viewing angles is essential, yet the methodologies to realize this goal within the 3DGS paradigm are currently underdeveloped.

The cornerstone of our approach lies in \emph{augmenting the interaction between Gaussians in both the in-painted and visible regions, encompassing geometry and appearance enhancements}. Initially, to bolster geometric consistency across the removal and visible areas, our method incorporates monocular depth estimation from multi-view images as a supplementary geometric constraint. This enhances the precision of 3D Gaussian Splatting (3DGS) placements. Employing a novel online depth alignment strategy, we refine the spatial arrangement of Gaussians within the removal area, ensuring improved alignment with adjacent regions. In terms of texture synthesis, our goal is to achieve a seamless blend between the visible and in-painted regions. Distinct from approaches tailored for implicit representations, which predominantly rely on image domain guidance for supervision, such as generating multi-view in-painted images~\cite{mirzaei2023spin,yin2023or,weder2023removing} or simulating pseudo-view-dependent effects from NeRF~\cite{mirzaei2023reference}, the explicit characteristic of Gaussian representations opens the door to innovative solutions. We introduce a novel method that facilitates feature interactions between Gaussian clusters from both visible and in-painted regions. This is achieved through a meticulously designed attention mechanism, which significantly improves the alignment of apparent and in-painted appearances. By sampling Gaussians positioned within both masked and unmasked areas, we refine their features via cross-attention in preparation for the final rendering. This self-interaction strategy capitalizes on the explicit nature of Gaussians to fine-tune the feature distribution in 3D spaces, culminating in enhanced coherence in the rendered outcomes. Furthermore, to mitigate the computational burden associated with directly manipulating millions of diminutive Gaussians, we implement a lightweight Gaussian Splatting architecture, Scaffold-GS~\cite{scaffoldgs}, as our base model. Scaffold-GS introduces a novel paradigm that organizes Gaussians around anchor points, using the features associated with these anchors to decode attributes for the respective Gaussians. This approach not only streamlines the processing of Gaussian data but also significantly enhances the efficiency and effectiveness of our rendering process.

To the end, we propose a holistic solution coined \textbf{GScream} for object \textbf{re}moval from \textbf{G}aussian \textbf{S}platting while maintaining the geo\textbf{m}etry and fe\textbf{a}ture \textbf{c}onsistency. The contribution of our paper is threefold summarized below:
\begin{itemize}
    \item We introduce GScream, a model that employs 3D Gaussian Splatting for object removal, specifically targeting and mitigating issues related to geometric inconsistencies and texture incoherence. This approach not only achieves significant efficiency but also ensures superior rendering quality when compared to traditional NeRF-based methods.
    \item To overcome the geometry inconsistency in the removal area, we incorporate multiview monocular depth estimation as an extra constraint. This aids in the precise optimization of Gaussian placements. Through an online depth alignment process, we enhance the geometric consistency between the removed area and the surrounding visible areas.
    \item Addressing the challenge of appearance incoherence, we exploit the explicit representation capability of 3DGS. We propose a unique feature regularization strategy that fosters improved interaction between Gaussian clusters in both the in-painted and visible sections of the scene. This method ensures coherence and elevates the appearance quality of the final rendered images.
\end{itemize}

% Related 
\section{Related Works}
\label{sec:related_works}
\subsection{Radiance Field for Novel View Synthesis}
Photo realistic view synthesis is a long-standing problem in computer vision and computer graphics~\cite{seitz1999photorealistic,kutulakos2000theory,sitzmann2019deepvoxels,neuralvolume}. Recently, the radiance field approaches~\cite{mildenhall2021nerf} revolutionized this task by only capturing scenes with multiple photos and brought the reconstruction quality to a new level with the help of neural implicit representations~\cite{sitzmann2019scene,park2019deepsdf} and effective positional encoding~\cite{mildenhall2021nerf,tancik2020fourier}. While the implicit representation benefits the optimization, the extensive queries of the network along the ray for rendering make the entire rendering speed costly and time-consuming~\cite{barron2022mip,barron2023zip}. Recently, there have been several attempts to facilitate the rendering speeds~\cite{chen2022mobilenerf,reiser2023merf,muller2022instant,3dgs}. Among all of them, the 3D Gaussian splitting (3DGS) representation~\cite{3dgs,scaffoldgs} stands as the most representative one which reaches a real-time rendering with state-of-the-art visual quality. 3DGS represents the radiance field as a collection of learnable 3D Gaussian. Each Gaussian blob includes information describing its 3D position, opacity, anisotropic covariance, and color features. With the dedicated design of a tiled-based splatting solution for training, the rendering of 3DGS is real-time with high quality. However, 3DGS is only proposed for novel view synthesis. It remains challenging to tame it if we want to remove objects from the pre-captured images.

\subsection{Object Removal from Radiance Field}
As the fidelity of 3D scene reconstruction advances, the ability to edit pre-captured 3D scenes becomes increasingly vital. Object removal, a key application in content generation, has garnered significant interest, particularly within the realm of radiance field representation. Several methods have been proposed to tackle this challenge~\cite{liu2022nerfin,weder2023removing,yin2023or,mirzaei2023spin,mirzaei2023reference}. For instance, NeRF-in~\cite{liu2022nerfin} and SPInNeRF~\cite{mirzaei2023spin} utilize 2D in-painting models to fill gaps in training views and rendered depths. However, these approaches often result in inconsistent in-painted images across different views, leading to ``ghost'' effects in the removed object regions. View-Subtitude~\cite{mirzaei2023reference} offers an alternative by in-painting a single reference image and designing depth-guided warping and bilateral filtering techniques to guide the generation in other views. 
Despite these innovations, the underlying issue of slow training and rendering speed persists in these NeRF-based methods. 
The recent 3DGS-based general editing framework, GaussianEditor~\cite{chen2023gaussianeditor}, includes the operation of deleting objects. However, despite its faster editing efficiency compared to NeRF-based methods, it still lacks specific constraints in the 3D domain. For the object removal task, purely fitting the 2D priors provided by the image in-painting model can also result in discontinuities in the 3D domain.

In response to these limitations, our work proposes a novel solution utilizing the 3D Gaussian Splatting (3DGS)~\cite{3dgs} representation to achieve efficient object removal. The 3DGS method offers a more rapid training and rendering process, making it a suitable candidate for this application. However, 3DGS, in its standard form, primarily focuses on RGB reconstruction loss, leading to less accurate underlying geometry for complex scenes. To make it suitable for recovering a scene without a selected object, we approach the problem in two stages: depth completion followed by texture propagation. We first enhance the geometric accuracy of 3DGS using monocular depth supervision. With a more refined geometric base, we then employ this improved structure to propagate 3D information outside the in-painted region to refine the texture in the in-painted region. These processes ensure not only the efficient removal of objects but also the maintenance of the scene's visual and geometric integrity.

% Method
\section{The Propose Framework: GScream}
\label{sec:method}

\begin{figure*}[!t]
    \centering
    \includegraphics[width=\linewidth]{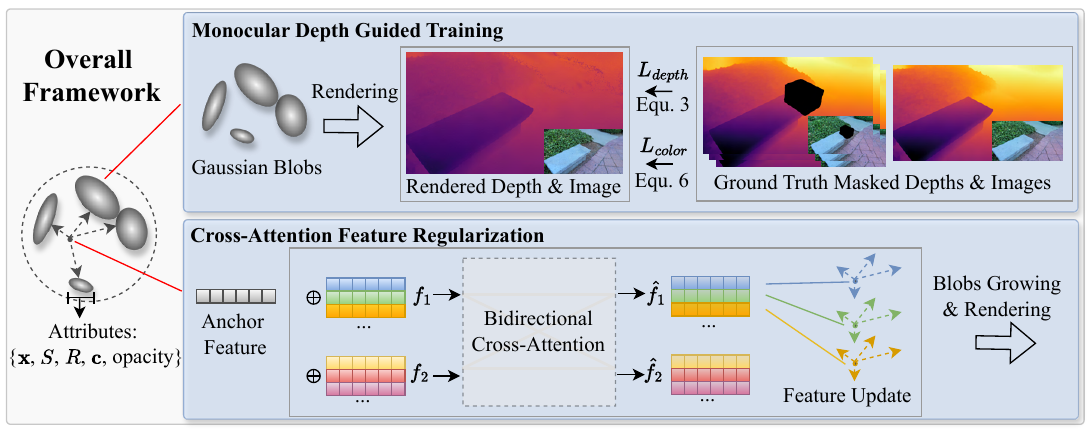}
    % \vspace{-10pt}
    \caption{\textbf{Illustration of our {GScream} framework.} It consists of two novel components, which are monocular depth guided training and cross-attention feature regularization. Our 3D Gaussian splatting (3DGS) representation is initialized by the 3D SfM points and supervised by both images and multi-view monocular depth estimation. The additional depth losses help refine the geometry accuracy within the 3DGS framework. The following 3D feature regularization performs texture propagation to refine the appearance within the 3D in-painted region. }
    % \vspace{-20pt}
    \label{fig:pipeline}
\end{figure*}

As illustrated in Fig.~\ref{fig:teaser}, given $N$ multi-view posed images $\left\{I_i| i=0, \dots, N\right\}$ %, $\left\{x_i, i \in[1, N]\right\}$
of a static real-world scene with the corresponding binary masks specifying the object $\left\{M_i|i=0, \dots, N\right\}$. The object mask $M_i$ is a binary mask with the object region set as $1$ and the background set as $0$. We assume these masks are provided for training, which can be obtained trivially by video segmentation~\cite{liu2022nerfin,cheng2021rethinking} or a straightforward 3D annotation~\cite{yin2023or,cen2023segment}. Our goal is to learn a 3D Gaussian representation to model the real-world scene with the object removed. To address this problem, we propose a novel framework named GScream, and the overview of it can be found in Fig.~\ref{fig:pipeline}. First, we select one view as the reference view and perform the 2D in-painting~\cite{Rombach_2022_CVPR,runwayml} to complete the content by the corresponding mask. Without loss of generality, we denote the selected view with index $0$ and the in-painted image as $\bar{I}$. We use the in-painted one single image to train the final 3DGS. The overview of our proposed GScream is shown in Fig.~\ref{fig:pipeline}.

The organization of this section is presented as follows: we will introduce the preliminary about 3D Gaussian Splatting and its variants in Sec.~\ref{sec:method_prelim}, and then dive into the details about the core design of our framework in terms of geometry consistency and appearance coherence in the following subsection.

\subsection{Preliminary: 3D Gaussain Splatting}\label{sec:method_prelim}

\noindent\textbf{3D Gaussian Splatting} We use the 3D Gaussian representation as our underlying modeling structure. Each Gaussian blob has the following attributes: 3D coordinates $\bm{\mu}$, scale matrix $S$, rotation matrix $R$, color features $\bm{c}$, and its opacity. 
With these attributes, the Gaussians are defined by the covariance matrix $\Sigma = RSS^TR^T$ centered at point $\bm{\mu}$:
\begin{equation}
    G(x) = \exp^{-\frac{1}{2}(\bm{x}-\bm{\mu})^T\Sigma^{-1}(\bm{x}-\bm{\mu})}.
\end{equation}
This Gaussian is multiplied by the opacity in the rendering process. By projecting the covariance onto the 2D plane following Zwicker~\etal~\cite{zwicker2001ewa}, we can obtain the projected Gaussian and adopt the volume rendering ($\alpha$-blending)~\cite{volume_rendering} to render the color in the image plane.
\begin{equation}
    \hat{C} = \sum_{k=1}^K \bm{c}_k\alpha_k\prod_{j=1}^{k-1}(1-\alpha_j),\label{eq:render_color}
\end{equation}
where $K$ means the number of sampling points along the ray and $\alpha$ is given by evaluating the projected Gaussian of $G(x)$ and the corresponding opacity. 
The initial 3D coordinates of each 3D Gaussian blob are initialized as the coordinates of the SfM points~\cite{schonberger2016structure}. All the attributes of Gaussians are optimized by the reconstruction loss of the image. More details can be found in~\cite{3dgs}.

\noindent\textbf{Scaffold-GS} While the sparse initial points are insufficient to model the entire scene, 3DGS designs a densification operation to split and merge Gaussians to capture more details. It will result in better rendering quality while leading to a heavy storage burden. Therefore, we adopt a lightweight Gaussian Splatting structure, Scaffold-GS~\cite{scaffoldgs}. The key contribution of it is to use anchors to generate new Gaussian attributes with several decoders. There will be a learnable feature embedding attached to each anchor, and all the new Gaussian attributes can be extracted from the anchor features. With the densification performed in the anchor points, the storage requirement of Scaffold-GS can be significantly reduced and benefit the modeling of the radiance field. We adopt it as our base model to propose an efficient object removal solution for Gaussian Splatting. More details can be found in~\cite{scaffoldgs}.

\subsection{Improve Geometry Consistency by Monocular Depth Guidance}\label{sec:depth_train}
One of the challenges to performing object removal upon 3DGS is the underlying geometry is too noisy~\cite{3dgs}, which further leads to difficulty when performing geometry completion for the removal region. To improve the quality, we propose to leverage the guidance from estimated monocular depth as extra supervision. Concretely, we use the depth estimation model~\cite{ke2023repurposing} to extract the depth $\mathcal{D}=\{D_i|i=0, \dots, N\}$ of each image from the in-painted image $\bar{I}$ and other views $\mathcal{I}$. Here $D_0$ corresponds to the estimated depth of $\bar{I}$.
% Online alignment

\noindent\textbf{Online Depth Alignment and Supervision}
The monocular depth estimation is not a metric depth~\cite{ke2023repurposing}. Therefore, we propose an online depth alignment design to utilize the depth guidance. However, the inconsistent depth estimation of $\bar{I}$ and $\mathcal{I}$ brings an additional issue. The $\mathcal{I}$ contains the object that we want to remove, while $\bar{I}$ depicts an image without the object. Therefore, we propose the following weighted depth loss to solve this problem:

% Depth loss 
%%%%%%%%%%%%%%%%%%%
\begin{equation}
\setlength{\abovedisplayskip}{6pt}
\setlength{\belowdisplayskip}{6pt}
\mathcal{L}_{\mathrm{depth}}=\frac{1}{HW}\sum M^{\prime}_i\|(w \hat{D}_i+q)-{D}_i\|,
\label{eq:depth_l1}
\end{equation}
%%%%%%%%%%%%%%%%%%%

%%%%%%%%%%%%%%%%%%%
\begin{equation}
\setlength{\abovedisplayskip}{6pt}
\setlength{\belowdisplayskip}{6pt}
M^{\prime}_i = \begin{cases} \lambda_1  M_i + \lambda_2  (1 - M_i), & \text { if } i=0 \\ \lambda_3   (1-M_i), & \text { if } i \neq 0\end{cases}.
\label{eq:depth_weight}
\end{equation}
%%%%%%%%%%%%%%%%%%% 
Where $\hat{D}_i$ is the rendered depth map from 3D Gaussian Splatting calculated similar to the Equ.~\ref{eq:render_color} by:
\begin{equation}
\hat{D} = \sum_{k=1}^K t_k\alpha_k\prod_{j=1}^{k-1}(1-\alpha_j),
\end{equation}
where $t_k$ is the 
z-coordinates of Gaussian mean $\mu_k$ in the corresponding camera coordinate system. The depth obtained from the monocular estimator $D_i$ and the rendered depth $\hat{D}_i$ by the 3D Gaussians have different numerical scales, so we cannot directly calculate the loss. We employ an online alignment method to address the scale issue. Specifically, we align the rendered depth using scale and shift parameters, denoted as $w$ and $q$, to match the scale of the monocular depth before calculating the loss. The scale and shift are obtained by solving a least-squares problem~\cite{ke2023repurposing,Yu2022MonoSDF}. For the image in $\mathcal{I}$, we only use the points outside the mask region, and the resulting scale and shift are applied to the entire depth map. We design different weights to calculate the depth loss as in Equ.~\ref{eq:depth_weight}. With this design, the depth supervision is applied to the entire depth map $D_0$ for the reference view, while it is applied on the background region for other views' depth $\{D_i|i=1, \dots, N\}$. The $\lambda_1, \lambda_2, \lambda_3$ are hyper-parameters to balance the influence of mask weights.
In addition to the point-wise L1 loss, we also enforce a total variation loss to enforce smoothness in the depth difference as follows: 
%%%%%%%%%%%%%%%%%%%
\begin{equation}
\setlength{\abovedisplayskip}{6pt}
\setlength{\belowdisplayskip}{6pt}
\mathcal{L}_{\mathrm{tv}}=\frac{1}{N}\sum M^{\prime}_i  \|\nabla((w \hat{D}_i+q)-{D}_i))\|
\label{eq:depth_smooth}
\end{equation}
%%%%%%%%%%%%%%%%%%%

\noindent\textbf{Color Loss} Following~\cite{3dgs,scaffoldgs}, we also apply the multi-view color reconstruction loss for both the training:
%%%%%%%%%%%%%%%%%%%
\begin{equation}
\begin{aligned}
\mathcal{L}_{\mathrm{color}} = \frac{1}{HW}\sum M^{\prime}_i &( (1-\lambda_{ssim})  \| \hat{C}_i - I_i \| \\
&+ \lambda_{ssim} SSIM(\hat{C}_i, I_i)),
\label{eq:color_loss} 
\end{aligned}
\end{equation}
where $\hat{C}$ is the rendered image from 3DGS. Thanks to the rendering efficiency of 3DGS, we can render the entire image and perform a structural image reconstruction loss~\cite{wang2003multiscale} SSIM to constrain the RGB image reconstruction.
%%%%%%%%%%%%%%%%%%% 
The overall training loss is the weighted sum of depth and color loss:
% Overall losses 
%%%%%%%%%%%%%%%%%%%
\begin{equation}
\setlength{\abovedisplayskip}{6pt}
\setlength{\belowdisplayskip}{6pt}
\begin{aligned}
\mathcal{L}_{\mathrm{total}} = \lambda_{depth} \mathcal{L}_{\mathrm{depth}} +\lambda_{tv} \mathcal{L}_{\mathrm{tv}} + \mathcal{L}_{\mathrm{color}}
\label{eq:total_loss} 
\end{aligned}
\end{equation}

\begin{figure}[!tp]
    \centering
    \includegraphics[width=\linewidth]{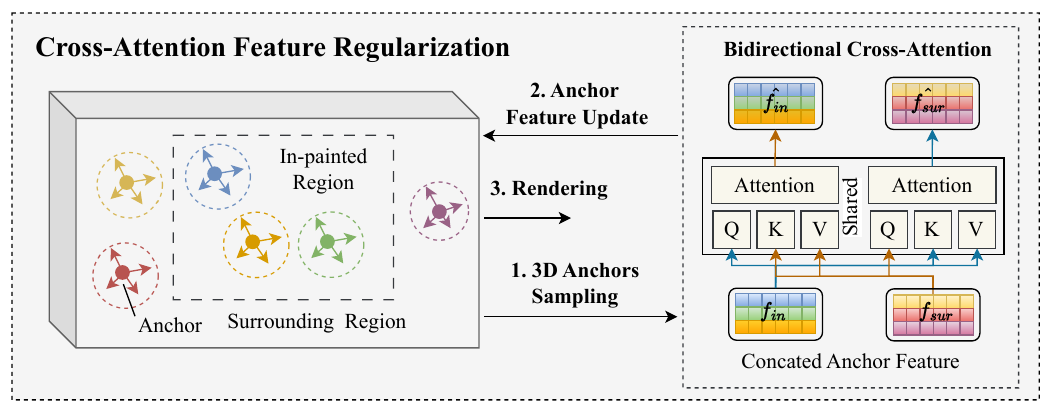}
    % \vspace{-15pt}
    \caption{\textbf{Illustration of the Cross-attention Feature Regularization.} Our regularization module consists of 3D Gaussian Sampling and a Bidirectional Cross-Attention Module, propagating the 3D feature from surrounding blobs to the in-painted region. As a complement to the 2D prior, the cross-attention mechanism enables the transmission of information among 3D Gaussian blobs, further ensuring the similarity of appearance between the in-painted region and its surroundings. }
    % \vspace{-18pt}
    \label{fig:propagation}
\end{figure}

%%%%%%%%%%%%%%%%%%%
\subsection{Cross-Attention Feature Regularization}\label{sec:prop}
% summary
Through monocular depth-guided training, we enhance the geometry of the 3D Gaussian representation. The following question is how we can refine the texture in the missing region from the surrounding environment.

Prior approaches in the realm of 3D object removal commonly employ a strategy that involves generating pseudo-RGB guidance to refresh the scene's information. This typically relies on leveraging multi-view in-painted images to update NeRF/3DGS models~\cite{mirzaei2023spin,yin2023or,chen2023gaussianeditor}, or on producing view-dependent effects as a form of guidance~\cite{mirzaei2023reference}. However, these methods tend to be sensitive to the quality of the pseudo-ground truth and often overlook the intrinsic relationships between the in-painted regions and their visible counterparts.

The key insight of our model is \emph{to propagate the accurate texture in the surrounding region into the in-painted region in a certain manner.} The explicit nature of 3DGS provides us the possibility to use the information from visible parts to update the content in the in-painted region. 
We expect this propagation can provide reliable information for the in-painted region in 3D space and ensure the propagated content is consistent across multiple viewpoints. Specifically, as shown in Fig.~\ref{fig:propagation}, we perform a two-stage procedure to achieve texture propagation, \emph{i.e.}, 3D anchors sampling, and subsequent bidirectional cross-attention. 

% 3D Gaussian sampling
\par\noindent\textbf{3D Gaussian Sampling} 
First, for each view $i$, we sample the patch that can simultaneously cover both the inside and the outside of the mask $M_i$. 
Then, we project the center coordinates of the 3D Gaussian anchors to the current view, to determine which anchor's 2D projection falls within the sampled 2D patch. 
After we identify the clusters of Gaussian anchors whose projections fall within the patch, we can easily categorize them into two groups based on whether their 2D projections are inside or outside the 2D mask. In this way, we sample 3D Gaussian anchors in both the in-painted and surrounding regions. 
Our goal is to sample 3D points in both the in-painted region and the surrounding region, as shown in the left part of Fig.~\ref{fig:propagation}. Although there are alternative sampling methods such as using depth for point back-projection, we believe that our approach based on 2D mask back-projection is sufficient to achieve our objectives.

% Bidirectional Cross-Attention
\par\noindent\textbf{Bidirectional Cross-Attention}
After obtaining the 3D Gaussian anchors from both regions, we perform bidirectional cross-attention between the two sets of Gaussian features to propagate information between the anchors. Specifically, we concatenate the two sets of Gaussian features as two tokens and take them as input to a bidirectional cross-attention structure following the classical definition~\cite{vaswani2017attention} $\text{Attention}(\mathbf{Q}, \mathbf{K}, \mathbf{V}) = \text{softmax}(\frac{\mathbf{Q}\mathbf{K}^{T}}{\sqrt{d_k}})\mathbf{V}$, where $d_k$ is the token length. 

The output of the cross-attention structure, which represents the updated features, is then assigned back to the corresponding Gaussian anchors. 
The bidirectional structure of the cross-attention is designed to facilitate bidirectional information propagation between the features inside and outside the in-painted regions. It can be seen as two sets of shared-parameter cross-attention modules, enabling information exchange between the two sets of features. As shown in Fig.~\ref{fig:propagation}, let us assume that the sampled tokens in the in-painted and surrounding regions are represented by the $f_{in}$ and $f_{sur}$. After passing them through the cross-attention module, the updated features can be denoted as $\hat{f}_{in}$ and $\hat{f}_{sur}$: 
\begin{equation}
\begin{aligned}
\hat{f}_{in}  &= \text{Attention}(\mathbf{Q}=f_{in}, \mathbf{K}=f_{sur}, \mathbf{V}=f_{sur}) \\
\hat{f}_{sur} &= \text{Attention}(\mathbf{Q}=f_{sur}, \mathbf{K}=f_{in}, \mathbf{V}=f_{in})
\end{aligned}
\label{eq:cross-attn} 
\end{equation}
As shown in Fig.~\ref{fig:pipeline}, when the sampled anchors complete the feature updates, all anchors undergo neural blobs growing and differentiable rendering as usual in~\cite{scaffoldgs}. The rendered depth map and image under the current viewpoint are then supervised by the total loss introduced in~\ref{eq:total_loss}. 

The 3D Gaussian sampling strategy together with the shared bidirectional cross-attention augments the anchor feature with similarity towards higher consistency. Through the gradients backpropagated to the anchors' features in the visible region, the similar anchors in the unpainted region can also be updated due to the attention mechanism. This design improves the consistency between the in-painted region and visible certain areas, which leads to better texture coherence in our experiments.

% Exp
\section{Experiments}
\label{sec:exp}

\subsection{Experimental Setup}
\par\noindent\textbf{Dataset} Following previous methods, we conducted experiments on the SPIn-NeRF dataset~\cite{mirzaei2023spin} and IBRNet dataset~\cite{wang2021ibrnet} for object removal. SPIn-NeRF dataset is proposed by~\cite{mirzaei2023spin}, consisting of 10 forward-facing in-the-wild scenes, including three indoor scenes and seven outdoor scenes. Each scene has 100 multi-view images with annotated foreground object masks. The training set consists of 60 images with objects, while the remaining 40 images without objects are used for testing. To ensure a fair comparison, we directly utilize the camera parameters and sparse reconstructed points from the dataset's released Structure-from-Motion (SfM) results instead of re-performing the sparse construction as~\cite{yin2023or}. IBRNet dataset is constructed by~\cite{wang2021ibrnet} for novel view synthesis, including selected scenes from existing datasets and 102 scenes collected by mobile phones. We use five real scenes captured by mobile phones from IBRNet for experimentation. More details can be found in the supplementary material.
\par\noindent\textbf{Baselines} We compare our methods with three recent baseline methods: SPIn-NeRF~\cite{mirzaei2023spin}, OR-NeRF~\cite{yin2023or}, and View-Sub~\cite{mirzaei2023reference}. We re-train and test the model using their open-source code to compare the first two baselines. We borrow the reported quantitative and qualitative results directly from the paper~\cite{mirzaei2023reference} due to the unavailable of open-source code. % during submission. 
% metrics
\par\noindent\textbf{Evaluation Metric} Since we are more interested in the visual quality for the final radiance field, we calculate the PSNR, SSIM~\cite{wang2004image}, and LPIPS~\cite{zhang2018unreasonable} scores on the full image and within the mask region. We also calculate the Frechet Inception Distance (FID)~\cite{heusel2017gans} score, which measures the similarity between the generated and real images regarding the feature distributions. Furthermore, we record the training time for each method to evaluate their efficiency. Please note that the scenes in IBRNet do not have ground truth images with objects removed, so quantitative metrics such as PSNR cannot be calculated. We only showcase partial quantitative results for these scenes. 
All experiments were conducted on an NVIDIA GeForce RTX 3090 with 24GB RAM. 

\subsection{Comparison with the State-of-the-art Methods} 
\begin{figure}[!t]
    \centering
    \includegraphics[width=\linewidth]{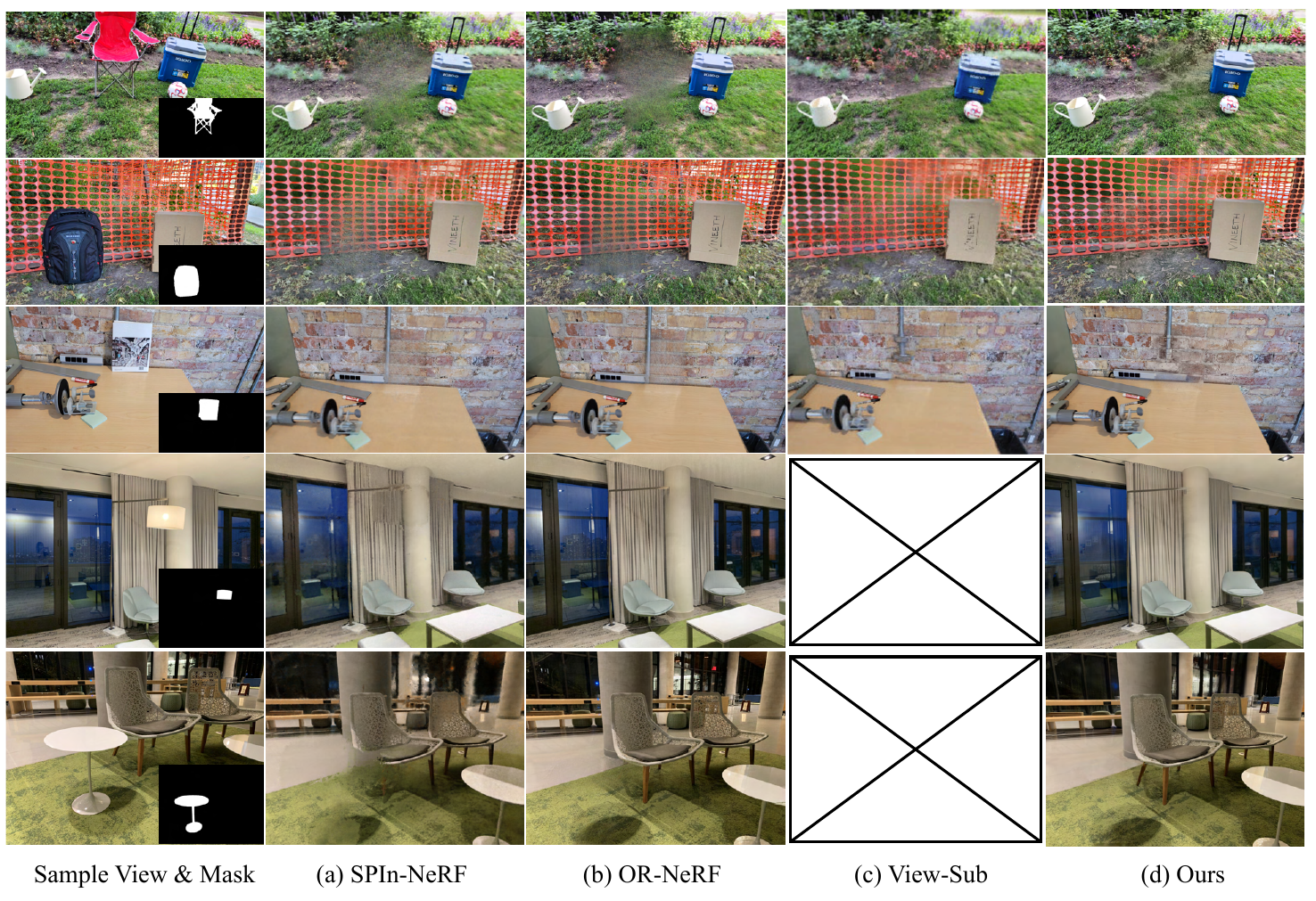}
    % \vspace{-15pt}
    \caption{\textbf{Qualitative results compared with the most representative object-removal approaches.} Illustration of the rendered qualitative images with object removed, compared with SPIn-NeRF~\cite{mirzaei2023spin}, OR-NeRF~\cite{yin2023or}, and View-Sub~\cite{mirzaei2023reference}. Our approach can synthesize high-quality images with natural removal effect.}
    % \vspace{-15pt}
    \label{fig:nvs}
\end{figure}

We present quantitative and qualitative comparisons between our method and three baseline methods in Tab.~\ref{tab:nvs} and Fig.~\ref{fig:nvs}, respectively. 

\par\noindent\textbf{Quantitative Comparison}
% table 1: metrics
As detailed in Tab.~\ref{tab:nvs}, our method either matches or surpasses SPIn-NeRF or OR-NeRF across all evaluated metrics. Notably, our approach yields superior similarity metrics, such as SSIM and LPIPS, suggesting that the images rendered by our method bear a closer resemblance to the ground truth in the test set. It is worth mentioning that SPIn-NeRF and OR-NeRF both utilize patch-based LPIPS loss in their optimization objective, which we did not employ. Despite this, our results still show an advantage in LPIPS, demonstrating the effectiveness of our method. Our method also performs better in terms of FID, indicating that the feature distribution of our rendered images is more consistent with real images without objects. 
Moreover, thanks to the efficiency of 3DGS representation rendering and optimization, our method achieves training times that are 1.5$\times$ and 4.0$\times$ faster than SPIn-NeRF and OR-NeRF, respectively. 
Regarding the View-sub method, due to the unavailability of its code, our comparison was limited to the masked LPIPS as reported in their paper, where our results were comparable. However, our method shows promise for an even more significant advantage in training efficiency.

%%%%%%%%%%%%%%%%%%%
\begin{table*}[t]
\caption{Quantitative comparison on novel view synthesis with the object removed. We compared our method with three baselines: SPIn-NeRF~\cite{mirzaei2023spin}, OR-NeRF~\cite{yin2023or}, and View-Sub~\cite{mirzaei2023reference}. ‘-’ indicates the metrics are not reported by the authors in the paper. ‘*’ indicates the metrics are directly borrowed from the paper of the corresponding method. }
\label{tab:nvs}
\resizebox{1.0\textwidth}{!}{
\begin{tabular}{l||ccccccccc}
\toprule[1.0pt]
\multicolumn{1}{c||}{Methods} & PSNR $\uparrow$ & masked PSNR $\uparrow$ & SSIM $\uparrow$ & masked SSIM $\uparrow$ & LPIPS$\downarrow$ & masked LPIPS $\downarrow$ & FID $\downarrow$ & Training Time $\downarrow$ \\ 
\midrule[1.0pt]
~ SPIn-NeRF~\cite{mirzaei2023spin}   & 20.18 & 15.80 & 0.46 & \pmb{0.21} & 0.47 & 0.58 & 58.78 & $\sim$ 3.0h \\
~ OR-NeRF~\cite{yin2023or}     & 20.32 & 15.74 & 0.54 & \pmb{0.21} & 0.35 & 0.56 & 38.69 & $\sim$ 6.0h \\
~ View-Sub~\cite{mirzaei2023reference}    & - & - & - & - & - & $\pmb{0.45}^{*}$ & - & - \\
\midrule[1.0pt]
~ GScream (Ours)   & \pmb{20.49} & \pmb{15.84} & \pmb{0.58} & \pmb{0.21} & \pmb{0.28} & 0.54 & \pmb{36.72} & $\sim$ \pmb{1.2h} \\
\bottomrule[1.0pt]
\end{tabular}
}

\end{table*}

%%%%%%%%%%%%%%%%%%%
%% Figure: NVS
%% Figure: NVS
%%%%%%%%%%%%%%%%%%%

\par\noindent\textbf{Qualitative Comparison} 
Fig.~\ref{fig:nvs} presents a qualitative comparison across five different scenes. For the first three scenes, we select of the nearest neighboring viewpoints based on the View-sub paper, enabling a coherent rendering comparison among all approaches. Despite slight camera pose differences, 
we believe these variations are negligible concerning the overall assessment of rendering quality. The leftmost column shows randomly selected scene images and their corresponding mask. 
% analysis 
Upon analysis, it is evident that while all methods exhibit competence in completing mask regions across certain scenarios, such as the regular wall depicted in the third row and the simple textured fence in the fourth row, SPIn-NeRF and OR-NeRF occasionally struggle with more complex regions. For instance, in the scenarios requiring the completion of both soil and bush textures (as seen in the first row), these methods often resort to inserting repetitive, unrealistic gray textures. In contrast, both the View-Sub method and our results can complete appropriate grass and plants. Similarly, in the second row, our completed railing appears more reasonable. 
While minor discrepancies in viewpoint exist between the results of the View-sub and ours, the fidelity of the completed textures in the first three scenes remains notably comparable.

Further analysis of the last two rows in Fig.~\ref{fig:nvs}, which shows two indoor scenes with more complex depth from the IBRNet dataset, reveals our method's proficiency. For instance, in the scenario involving lamp removal, our method naturally completes the curtain behind the lamp compared to SPIn-Nerf and OR-Nerf. In the case of table removal, our method reconstructs the chair legs and carpet more accurately.

\subsection{Ablation Study} 
\begin{figure}[!t]
    \centering
    \includegraphics[width=0.9\linewidth]{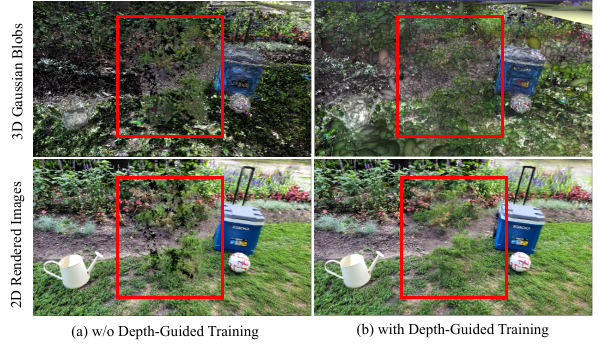}
    % \vspace{-10pt}
    \caption{\textbf{Qualitative results of the effective depth-guided training.} We visualize the scene in 3D Gaussian Splatting format and 2D rendered image by ablating the depth-guide training. The geometry guidance provides more information to fill the missing area with Gaussian blobs.}
    % \vspace{-20pt} 
    \label{fig:abl_3d} 
\end{figure}

\begin{figure*}[!t]
    \centering
    \includegraphics[width=\linewidth]{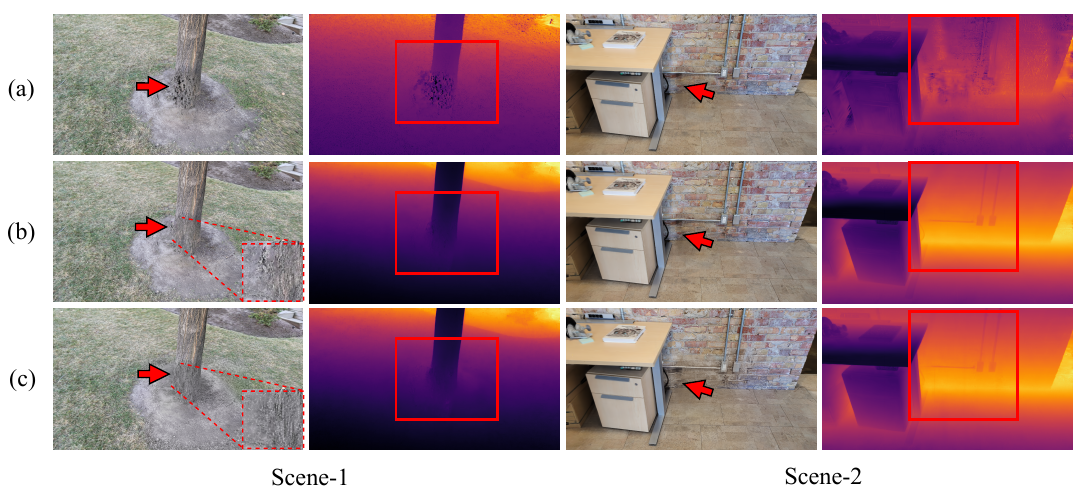}
    % \vspace{-20pt}
    \caption{\textbf{Qualitative results of the ablation study.} We provide the visualization of different variants of our method. From the top to bottom, \textbf{(a)} 
    GScream w/o Cross-Attention \& Mono-Depth; \textbf{(b)} GScream w/o Cross-Attention Feature Regularization; \textbf{(c)} Our Full Method GScream. We visualize the rendered RGB and depth to verify the effectiveness of our proposed components. Our full model produces a more reasonable depth and RGB image. Please zoom in for a better view. }
    % \vspace{-21pt}
    \label{fig:abl_2d}
\end{figure*}

We conduct ablation experiments on mono-depth supervision and cross-attention feature regularization and present the quantitative and qualitative results in Tab.~\ref{tab:abl}, Fig.~\ref{fig:abl_3d}, and Fig.~\ref{fig:abl_2d}.

\par\noindent\textbf{Analysis of Depth Supervision}
We first analyze our first contribution: introducing multi-view depth maps to aid in the 3D geometry learning of the in-painted area. Fig.~\ref{fig:abl_3d} (a) and (b) show the results supervised by using Equ.~\ref{eq:color_loss} and Equ.~\ref{eq:total_loss} based on the original Scaffold-GS. Note that the former does not have mono-depth supervision while the latter has mono-depth supervision. 
We visualize the learned Gaussian blobs and 2D images before and after incorporating depth supervision (all visualized in novel views). From Fig.~\ref{fig:abl_3d}, we can observe that in (a), where depth supervision is lacking, the positions of the Gaussian blobs within the red box are floating in the air, with noticeable holes interspersed in between. The corresponding 2D rendered images also exhibit noticeable texture floating. 
However, the involvement of depth supervision in (b) leads to more plausible positions of the 3D Gaussian blobs: most blobs are located within areas with objects (grass and bushes) rather than floating in the air as in (a). The corresponding 2D rendered images are noticeably more realistic and plausible. 
This demonstrates that our depth supervision significantly constrains the position of Gaussian blobs and improves the geometric accuracy of 3DGS, which enables the realism of the 2D renderings in novel views.

\par\noindent\textbf{Quantitative Analysis of Key Components} 
We further disable Mono-Depth Supervision and Cross-Attention Feature Regularization modules individually based on the full model GScream, and present more quantitative and qualitative results of these ablation experiments in Tab.~\ref{tab:abl} and Fig.~\ref{fig:abl_2d}. 
Disabling Cross-Attention Feature Regularization means training only with Equ.~\ref{eq:total_loss}, without performing 3D Gaussian sampling and bidirectional cross-attention. 
Disabling both means only retaining the color loss term in Equ.~\ref{eq:total_loss}.

%%%%%%%%%%%%%%%%%%%
\begin{table*}[t]
\caption{\textbf{Quantitative comparison of different variants of our proposed method.} We remove one or both of the Mono-Depth Supervision and Cross-Attn (Cross-Attention) Regularization components and compare the quantitative results.}
\label{tab:abl}
% \vspace{-8pt}
\resizebox{1.0\textwidth}{!}{
\begin{tabular}{l||ccccccccc}
\toprule
\multicolumn{1}{c||}{Variants} & PSNR $\uparrow$ & masked-PSNR $\uparrow$ & SSIM $\uparrow$ & masked-SSIM $\uparrow$ & LPIPS$\downarrow$ & masked-LPIPS $\downarrow$  \\ 
\midrule[1.0pt]
~ GScream w/o Cross-Attn \& Mono-Depth & 20.12 & 14.87 & 0.58 & 0.19 & \pmb{0.26} & 0.56 \\
~ GScream w/o Cross-Attn & 20.47 & 15.63 & 0.58 & 0.20 & \pmb{0.26} & \pmb{0.50} \\
\midrule[1.0pt]
~ GScream (Our Full Model)    & \pmb{20.49} & \pmb{15.84} & \pmb{0.58} & \pmb{0.21} & 0.28 & 0.54 \\
\bottomrule[1.0pt]
\end{tabular}
}
\vspace{-15pt}
\end{table*}
%%%%%%%%%%%%%%%%%%%

From the Tab.~\ref{tab:abl}, we can observe that removing the cross-attention feature regularization modules leads to a degradation in the metrics PSNR and SSIM. For instance, the masked PSNR decreases from 15.84 to 15.63, 
indicating that the content filled in the masked regions becomes less reasonable. 
This suggests that improvements in depth accuracy and feature propagation are beneficial for the results. 
Furthermore, if both modules are disabled, the metrics become even worse. Compared to the full model, the masked PSNR decreases to 14.87, and the masked-SSIM further decreases to 0.19, suggesting poorer depth and no 3D regularization in masked regions lead to worse results. 

% figure 
\par\noindent\textbf{Qualitative Analysis of the Mono-Depth Module.}
The label (a)(b)(c) in Fig.~\ref{fig:abl_2d} represent (a) GScream w/o Cross-Attention \& Mono-Depth; (b) GScream w/o Cross-Attention Regularization and (c) Our Full Method (GScream), respectively. 
For both Scene-1 and Scene-2, by comparing Fig.~\ref{fig:abl_2d} (a) with (b)(c), we can observe that removing depth supervision results in poor depth prediction, with significant noise present in the red box region and along the image edges. The texture quality of scene (a) suffers notably due to the absence of depth supervision, resulting in texture holes when viewed from novel perspectives. 

\par\noindent\textbf{Qualitative Analysis of the Cross-Attention Module.} 
While our experiments revealed a marginal reduction in the LPIPS metric upon deactivating the cross-attention module, we are poised to showcase this module's substantial role in enhancing our results in Fig.~\ref{fig:abl_2d}. 
While Fig.~\ref{fig:abl_2d} (b) benefits from the incorporation of monocular depth supervision, leading to improved texture filling and depth accuracy, the outcomes still fall short of naturalness due to the absence of 3D feature regularization. 
In Scene-1 (b) of Fig.~\ref{fig:abl_2d}, when the perspective shifts to the left side of the tree trunk, black holes become visible in areas distanced from the frontal view, as indicated by the red arrow (zooming in is recommended for clarity). This scenario underscores the limitations of solely relying on 2D priors for supervision, which are unable to remediate texture gaps in unseen regions. However, the introduction of 3D feature regularization in (c) effectively addresses these shortcomings by filling the previously observed holes. This enhancement reveals the critical role of 3D feature interactions in supplementing 2D priors, enabling the propagation of appropriate textures to obscured areas and thereby ensuring more cohesive rendering in novel views.
In Scene-2, a side-by-side comparison of (a) and (b) reveals that, while (b) demonstrates depth enhancements over (a), both still exhibit a pronounced sharp boundary, as indicated by the red arrow, which detracts from naturalism. However, integrating feature cross-attention in (c) significantly mitigates this issue. The previously stark gap softens, eliminating the noticeable boundary. This transformation suggests that facilitating feature information exchange can harmonize originally disjointed textures at boundaries, ensuring a more seamless and consistent texture transition.

%%%%%%%%%%%%%%%%%%%
%% Figure: ABLATION 2D
%% Figure: ABLATION 2D
%%%%%%%%%%%%%%%%%%%

% Conclusion
\section{Conclusion}
\label{sec:conclusion}

% Conclusion 
In conclusion, our innovative framework for object removal, which leverages 3D Gaussian Splatting, has proven to be both effective and more efficient than traditional NeRF-based approaches. Through the integration of monocular-depth guided training and cross-attention feature regularization techniques, our method facilitates rapid training speeds while simultaneously preserving multi-view geometric and texture consistency in the inpainted textures. Experimental validations confirm that our approach outperforms existing NeRF-based methods in terms of both efficiency and effectiveness.

% \clearpage
% \appendix
\section{Supplementary Material}
\label{sec:supp}

In this supplementary material, we provide additional experimental results in Sec.~\ref{sec:supp_exp}. We also provide a video demo for more qualitative state-of-the-art comparisons and ablation studies as discussed in Sec.~\ref{sec:supp_video}.

\section{Additional Experiments}\label{sec:supp_exp}

\subsection{Comparison with GaussianEditor}
\begin{figure}[htbp]
    \centering
    \includegraphics[width=1.0\linewidth]{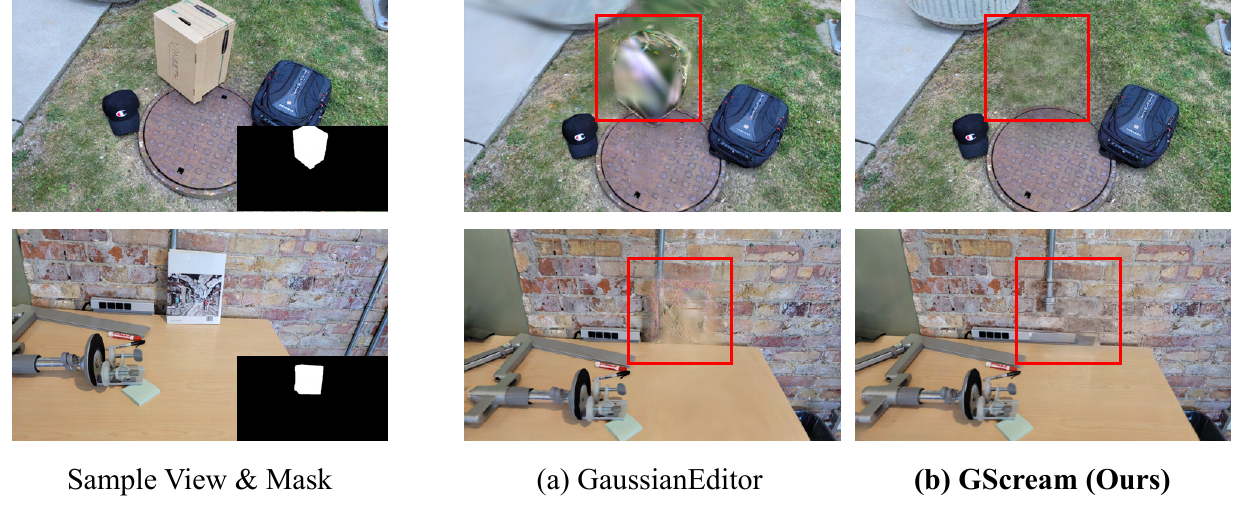}
    % \vspace{-15pt}
    \caption{\textbf{Qualitative comparison of object removal with the 3D Gaussian Splatting-based method GaussianEditor~\cite{chen2023gaussianeditor}.} }
    % \vspace{-20pt} 
    \label{fig:supp_ge} 
\end{figure}

In addition to the comparisons in the main text with the NeRF-based method~\cite{mirzaei2023spin,yin2023or,mirzaei2023reference}, we also compared our approach \textbf{GScream} with the GaussianEditor~\cite{chen2023gaussianeditor}, which is a general editing pipeline also based on Gaussian Splatting representation. 
They utilize 2D prior from the diffusion model to guide the updates of the Hierarchical Gaussian splatting (HGS) in order to achieve stabilized editing. 
The qualitative comparison of object removal with GaussianEditor is illustrated in Fig.~\ref{fig:supp_ge}. From the Fig.~\ref{fig:supp_ge}, our method can fill the mask region with more realistic and plausible textures.

% figure 1: supp_ge

\subsection{Ablations on using different Depth Estimation Models}
\begin{figure}[htbp]
    \centering
    \includegraphics[width=1.0\linewidth]{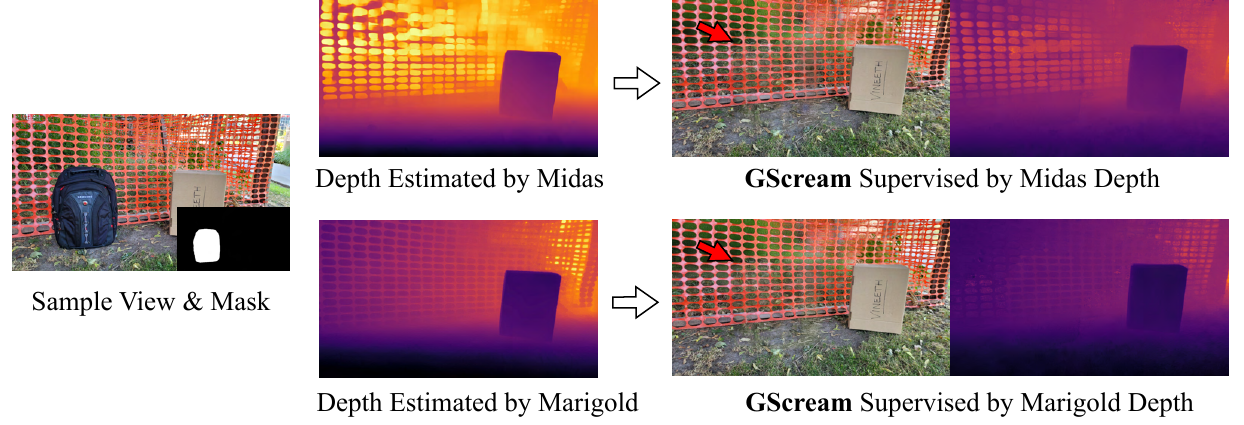}
    % \vspace{-15pt}
    \caption{\textbf{Qualitative results of using different monocular depth estimation models: Midas~\cite{birkl2023midas} and Marigold~\cite{ke2023repurposing}. } }
    % \vspace{-20pt} 
    \label{fig:supp_depth} 
\end{figure}

In our method GScream, we use a depth estimation model to obtain depth maps for each viewpoint independently. In this subsection, we compare two different single image depth estimation methods: Midas~\cite{birkl2023midas} and Marigold~\cite{ke2023repurposing} for obtaining depth prior, which is utilized to supervise our GScream. As shown in Fig.~\ref{fig:supp_depth}, the depth predicted by Midas at the red fence is not particularly continuous, resulting in the texture and depth of the learned GScream at the red fence being less continuous as well. On the other hand, Marigold can predict more continuous depth, thereby guiding a relatively more continuous GScream. This also demonstrates that accurate depth guidance is crucial for learning geometric continuity in the representation of 3D Gaussian Splatting. 

% figure 2: abl: depth model

% figure 3: abl: in-painting model
\begin{figure}[htbp]
    \centering
    \includegraphics[width=1.0\linewidth]{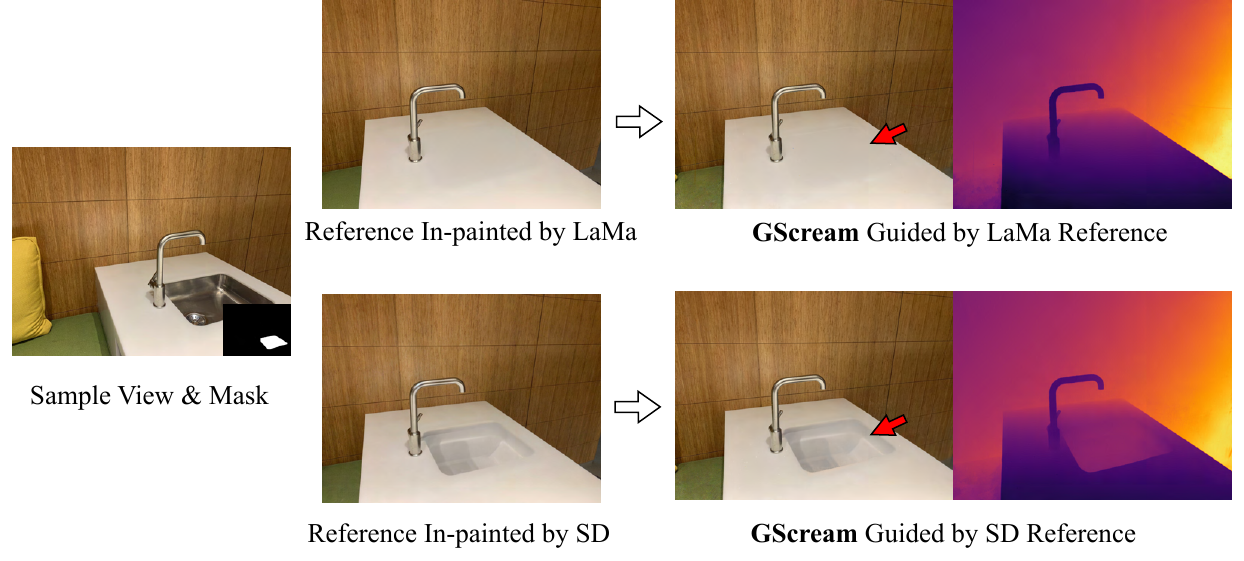}
    % \vspace{-15pt}
    \caption{\textbf{Qualitative results of using different 2D in-painting models: LaMa~\cite{suvorov2022resolution} and Stable Diffusion (SD)~\cite{runwayml}} }
    % \vspace{-20pt} 
    \label{fig:supp_inpaint} 
\end{figure}

\subsection{Ablations on using different 2D In-Painting Models}
% analysis
In our method, we use a 2D in-painting model to obtain a reference image from a certain viewpoint as guidance. In this subsection, we compare two different methods using different 2D in-painting models: LaMa~\cite{suvorov2022resolution} and Stable Diffusion~\cite{runwayml} to obtain guidance for GScream. As shown in Fig.~\ref{fig:supp_inpaint}, LaMa removes both the metal sink and the indentation, while Stable Diffusion removes the metal sink but retains the indentation. Both results from these two in-painting methods can serve as guidance for obtaining reasonable GScream results. 
This indicates that the choice of the in-painting method for obtaining the reference is not crucial; what matters more is obtaining a reasonable reference. As long as there is a reasonable reference, GScream can generate 3D geometry and texture continuous results.

\section{Video Demo}\label{sec:supp_video}
% video 
We also provide a demo video for more qualitative state-of-the-art comparisons and ablation studies. The video demo can be found from the project page: \url{https://w-ted.github.io/publications/gscream}

% \section*{References}

% References follow the acknowledgments in the camera-ready paper. Use unnumbered first-level heading for
% the references. Any choice of citation style is acceptable as long as you are
% consistent. It is permissible to reduce the font size to \verb+small+ (9 point)
% when listing the references.
% Note that the Reference section does not count towards the page limit.
% \medskip

% {
% \small
% \input{main.bbl}

% [1] Alexander, J.A.\ \& Mozer, M.C.\ (1995) Template-based algorithms for
% connectionist rule extraction. In G.\ Tesauro, D.S.\ Touretzky and T.K.\ Leen
% (eds.), {\it Advances in Neural Information Processing Systems 7},
% pp.\ 609--616. Cambridge, MA: MIT Press.

% [2] Bower, J.M.\ \& Beeman, D.\ (1995) {\it The Book of GENESIS: Exploring
%   Realistic Neural Models with the GEneral NEural SImulation System.}  New York:
% TELOS/Springer--Verlag.

% [3] Hasselmo, M.E., Schnell, E.\ \& Barkai, E.\ (1995) Dynamics of learning and
% recall at excitatory recurrent synapses and cholinergic modulation in rat
% hippocampal region CA3. {\it Journal of Neuroscience} {\bf 15}(7):5249-5262.
% }

%%%%%%%%%%%%%%%%%%%%%%%%%%%%%%%%%%%%%%%%%%%%%%%%%%%%%%%%%%%%

% {
% \balance
% \small
% \bibliographystyle{ieee_fullname}
% \bibliography{main}

\begin{thebibliography}{10}\itemsep=-1pt

\bibitem{barron2022mip}
Jonathan~T Barron, Ben Mildenhall, Dor Verbin, Pratul~P Srinivasan, and Peter Hedman.
\newblock Mip-nerf 360: Unbounded anti-aliased neural radiance fields.
\newblock In {\em CVPR}, 2022.

\bibitem{barron2023zip}
Jonathan~T Barron, Ben Mildenhall, Dor Verbin, Pratul~P Srinivasan, and Peter Hedman.
\newblock Zip-nerf: Anti-aliased grid-based neural radiance fields.
\newblock {\em arXiv preprint arXiv:2304.06706}, 2023.

\bibitem{bertalmio2000image}
Marcelo Bertalmio, Guillermo Sapiro, Vincent Caselles, and Coloma Ballester.
\newblock Image inpainting.
\newblock In {\em Proceedings of the 27th annual conference on Computer graphics and interactive techniques}, pages 417--424, 2000.

\bibitem{birkl2023midas}
Reiner Birkl, Diana Wofk, and Matthias M{\"u}ller.
\newblock Midas v3.1 -- a model zoo for robust monocular relative depth estimation.
\newblock {\em arXiv preprint arXiv:2307.14460}, 2023.

\bibitem{cen2023segment}
Jiazhong Cen, Zanwei Zhou, Jiemin Fang, Wei Shen, Lingxi Xie, Xiaopeng Zhang, and Qi Tian.
\newblock Segment anything in 3d with nerfs.
\newblock {\em arXiv preprint arXiv:2304.12308}, 2023.

\bibitem{chen2023gaussianeditor}
Yiwen Chen, Zilong Chen, Chi Zhang, Feng Wang, Xiaofeng Yang, Yikai Wang, Zhongang Cai, Lei Yang, Huaping Liu, and Guosheng Lin.
\newblock Gaussianeditor: Swift and controllable 3d editing with gaussian splatting.
\newblock In {\em CVPR}, 2024.

\bibitem{chen2022mobilenerf}
Zhiqin Chen, Thomas Funkhouser, Peter Hedman, and Andrea Tagliasacchi.
\newblock Mobilenerf: Exploiting the polygon rasterization pipeline for efficient neural field rendering on mobile architectures.
\newblock In {\em CVPR}, 2023.

\bibitem{cheng2021rethinking}
Ho~Kei Cheng, Yu-Wing Tai, and Chi-Keung Tang.
\newblock Rethinking space-time networks with improved memory coverage for efficient video object segmentation.
\newblock In {\em NeurIPS}, 2021.

\bibitem{dai2020sg}
Angela Dai, Christian Diller, and Matthias Nie{\ss}ner.
\newblock Sg-nn: Sparse generative neural networks for self-supervised scene completion of rgb-d scans.
\newblock In {\em CVPR}, 2020.

\bibitem{dai2017complete}
Angela Dai, Charles~Ruizhongtai Qi, and Matthias Nie{\ss}ner.
\newblock Shape completion using 3d-encoder-predictor cnns and shape synthesis.
\newblock In {\em CVPR}, 2017.

\bibitem{dai2018scancomplete}
Angela Dai, Daniel Ritchie, Martin Bokeloh, Scott Reed, J{\"u}rgen Sturm, and Matthias Nie{\ss}ner.
\newblock Scancomplete: Large-scale scene completion and semantic segmentation for 3d scans.
\newblock In {\em CVPR}, 2018.

\bibitem{heusel2017gans}
Martin Heusel, Hubert Ramsauer, Thomas Unterthiner, Bernhard Nessler, and Sepp Hochreiter.
\newblock Gans trained by a two time-scale update rule converge to a local nash equilibrium.
\newblock In {\em NeurIPS}, 2017.

\bibitem{kazhdan2006poisson}
Michael Kazhdan, Matthew Bolitho, and Hugues Hoppe.
\newblock Poisson surface reconstruction.
\newblock In {\em Proceedings of the fourth Eurographics symposium on Geometry processing}, volume~7, page~0, 2006.

\bibitem{ke2023repurposing}
Bingxin Ke, Anton Obukhov, Shengyu Huang, Nando Metzger, Rodrigo~Caye Daudt, and Konrad Schindler.
\newblock Repurposing diffusion-based image generators for monocular depth estimation.
\newblock In {\em CVPR}, 2024.

\bibitem{3dgs}
Bernhard Kerbl, Georgios Kopanas, Thomas Leimk{\"u}hler, and George Drettakis.
\newblock 3d gaussian splatting for real-time radiance field rendering.
\newblock {\em ToG}, 42(4):1--14, 2023.

\bibitem{kutulakos2000theory}
Kiriakos~N Kutulakos and Steven~M Seitz.
\newblock A theory of shape by space carving.
\newblock {\em IJCV}, 38:199--218, 2000.

\bibitem{liu2022nerfin}
Hao-Kang Liu, I Shen, Bing-Yu Chen, et~al.
\newblock Nerf-in: Free-form nerf inpainting with rgb-d priors.
\newblock {\em arXiv preprint arXiv:2206.04901}, 2022.

\bibitem{neuralvolume}
Stephen Lombardi, Tomas Simon, Jason Saragih, Gabriel Schwartz, Andreas Lehrmann, and Yaser Sheikh.
\newblock Neural volumes: Learning dynamic renderable volumes from images.
\newblock {\em ACM Trans. Graph.}, 38(4):65:1--65:14, July 2019.

\bibitem{scaffoldgs}
Tao Lu, Mulin Yu, Linning Xu, Yuanbo Xiangli, Limin Wang, Dahua Lin, and Bo Dai.
\newblock Scaffold-gs: Structured 3d gaussians for view-adaptive rendering.
\newblock In {\em CVPR}, 2024.

\bibitem{volume_rendering}
Nelson Max.
\newblock Optical models for direct volume rendering.
\newblock {\em TVCG}, 1(2):99--108, 1995.

\bibitem{mildenhall2021nerf}
Ben Mildenhall, Pratul~P Srinivasan, Matthew Tancik, Jonathan~T Barron, Ravi Ramamoorthi, and Ren Ng.
\newblock Nerf: Representing scenes as neural radiance fields for view synthesis.
\newblock {\em Communications of the ACM}, 65(1):99--106, 2021.

\bibitem{mirzaei2023reference}
Ashkan Mirzaei, Tristan Aumentado-Armstrong, Marcus~A Brubaker, Jonathan Kelly, Alex Levinshtein, Konstantinos~G Derpanis, and Igor Gilitschenski.
\newblock Reference-guided controllable inpainting of neural radiance fields.
\newblock In {\em ICCV}, 2023.

\bibitem{mirzaei2023spin}
Ashkan Mirzaei, Tristan Aumentado-Armstrong, Konstantinos~G Derpanis, Jonathan Kelly, Marcus~A Brubaker, Igor Gilitschenski, and Alex Levinshtein.
\newblock Spin-nerf: Multiview segmentation and perceptual inpainting with neural radiance fields.
\newblock In {\em CVPR}, 2023.

\bibitem{muller2022instant}
Thomas M{\"u}ller, Alex Evans, Christoph Schied, and Alexander Keller.
\newblock Instant neural graphics primitives with a multiresolution hash encoding.
\newblock {\em ToG}, 41(4):1--15, 2022.

\bibitem{park2019deepsdf}
Jeong~Joon Park, Peter Florence, Julian Straub, Richard Newcombe, and Steven Lovegrove.
\newblock Deepsdf: Learning continuous signed distance functions for shape representation.
\newblock In {\em CVPR}, 2019.

\bibitem{reiser2023merf}
Christian Reiser, Richard Szeliski, Dor Verbin, Pratul~P Srinivasan, Ben Mildenhall, Andreas Geiger, Jonathan~T Barron, and Peter Hedman.
\newblock Merf: Memory-efficient radiance fields for real-time view synthesis in unbounded scenes.
\newblock {\em arXiv preprint arXiv:2302.12249}, 2023.

\bibitem{Rombach_2022_CVPR}
Robin Rombach, Andreas Blattmann, Dominik Lorenz, Patrick Esser, and Bj\"orn Ommer.
\newblock High-resolution image synthesis with latent diffusion models.
\newblock In {\em CVPR}, 2022.

\bibitem{runwayml}
RunwayML.
\newblock Stable diffusion.
\newblock \url{https://huggingface.co/runwayml/stable-diffusion-inpainting}, 2021.

\bibitem{schonberger2016structure}
Johannes~L Schonberger and Jan-Michael Frahm.
\newblock Structure-from-motion revisited.
\newblock In {\em CVPR}, 2016.

\bibitem{seitz1999photorealistic}
Steven~M Seitz and Charles~R Dyer.
\newblock Photorealistic scene reconstruction by voxel coloring.
\newblock {\em IJCV}, 35:151--173, 1999.

\bibitem{shih20203d}
Meng-Li Shih, Shih-Yang Su, Johannes Kopf, and Jia-Bin Huang.
\newblock 3d photography using context-aware layered depth inpainting.
\newblock In {\em CVPR}, 2020.

\bibitem{sitzmann2019deepvoxels}
Vincent Sitzmann, Justus Thies, Felix Heide, Matthias Nie{\ss}ner, Gordon Wetzstein, and Michael Zollhofer.
\newblock Deepvoxels: Learning persistent 3d feature embeddings.
\newblock In {\em CVPR}, 2019.

\bibitem{sitzmann2019scene}
Vincent Sitzmann, Michael Zollh{\"o}fer, and Gordon Wetzstein.
\newblock Scene representation networks: Continuous 3d-structure-aware neural scene representations.
\newblock In {\em NeurIPS}, 2019.

\bibitem{suvorov2022resolution}
Roman Suvorov, Elizaveta Logacheva, Anton Mashikhin, Anastasia Remizova, Arsenii Ashukha, Aleksei Silvestrov, Naejin Kong, Harshith Goka, Kiwoong Park, and Victor Lempitsky.
\newblock Resolution-robust large mask inpainting with fourier convolutions.
\newblock In {\em WACV}, 2022.

\bibitem{tancik2020fourier}
Matthew Tancik, Pratul Srinivasan, Ben Mildenhall, Sara Fridovich-Keil, Nithin Raghavan, Utkarsh Singhal, Ravi Ramamoorthi, Jonathan Barron, and Ren Ng.
\newblock Fourier features let networks learn high frequency functions in low dimensional domains.
\newblock In {\em NeurIPS}, 2020.

\bibitem{vaswani2017attention}
Ashish Vaswani, Noam Shazeer, Niki Parmar, Jakob Uszkoreit, Llion Jones, Aidan~N Gomez, {\L}ukasz Kaiser, and Illia Polosukhin.
\newblock Attention is all you need.
\newblock In {\em NeurIPS}, pages 5998--6008, 2017.

\bibitem{wang2023inpaintnerf360}
Dongqing Wang, Tong Zhang, Alaa Abboud, and Sabine S{\"u}sstrunk.
\newblock Inpaintnerf360: Text-guided 3d inpainting on unbounded neural radiance fields.
\newblock {\em arXiv preprint arXiv:2305.15094}, 2023.

\bibitem{wang2021ibrnet}
Qianqian Wang, Zhicheng Wang, Kyle Genova, Pratul Srinivasan, Howard Zhou, Jonathan~T. Barron, Ricardo Martin-Brualla, Noah Snavely, and Thomas Funkhouser.
\newblock Ibrnet: Learning multi-view image-based rendering.
\newblock In {\em CVPR}, 2021.

\bibitem{wang2004image}
Zhou Wang, Alan~C Bovik, Hamid~R Sheikh, and Eero~P Simoncelli.
\newblock Image quality assessment: from error visibility to structural similarity.
\newblock {\em TIP}, 13(4):600--612, 2004.

\bibitem{wang2003multiscale}
Zhou Wang, Eero~P Simoncelli, and Alan~C Bovik.
\newblock Multiscale structural similarity for image quality assessment.
\newblock In {\em The Thrity-Seventh Asilomar Conference on Signals, Systems \& Computers, 2003}, volume~2, pages 1398--1402. Ieee, 2003.

\bibitem{weder2023removing}
Silvan Weder, Guillermo Garcia-Hernando, Aron Monszpart, Marc Pollefeys, Gabriel~J Brostow, Michael Firman, and Sara Vicente.
\newblock Removing objects from neural radiance fields.
\newblock In {\em CVPR}, 2023.

\bibitem{yin2023or}
Youtan Yin, Zhoujie Fu, Fan Yang, and Guosheng Lin.
\newblock Or-nerf: Object removing from 3d scenes guided by multiview segmentation with neural radiance fields.
\newblock {\em arXiv preprint arXiv:2305.10503}, 2023.

\bibitem{Yu2022MonoSDF}
Zehao Yu, Songyou Peng, Michael Niemeyer, Torsten Sattler, and Andreas Geiger.
\newblock Monosdf: Exploring monocular geometric cues for neural implicit surface reconstruction.
\newblock In {\em NeurIPS}, 2022.

\bibitem{zhang2018unreasonable}
Richard Zhang, Phillip Isola, Alexei~A Efros, Eli Shechtman, and Oliver Wang.
\newblock The unreasonable effectiveness of deep features as a perceptual metric.
\newblock In {\em CVPR}, 2018.

\bibitem{zwicker2001ewa}
Matthias Zwicker, Hanspeter Pfister, Jeroen Van~Baar, and Markus Gross.
\newblock Ewa volume splatting.
\newblock In {\em VIS}, 2001.

\end{thebibliography}
% }

\end{document}